\documentclass[10pt, a4paper]{article}
\usepackage[]{lrec-coling2024}

\usepackage{graphicx}
\usepackage{tabularx}
\usepackage{soul}
\usepackage{float}
\usepackage{comment}
\usepackage{subcaption}

\usepackage{xcolor}
\usepackage{hyperref}
 \definecolor{darkblue}{rgb}{0, 0, 0.5}
  \hypersetup{colorlinks=true, citecolor=darkblue, linkcolor=darkblue, urlcolor=darkblue}

\usepackage{xstring}

\usepackage{color}

\usepackage{todonotes}
\usepackage{booktabs}
\usepackage[british]{babel}
\definecolor{cadmiumgreen}{rgb}{0.0, 0.42, 0.24}
\definecolor{cadmiumred}{rgb}{0.89, 0.0, 0.13}
\definecolor{cadmiumorange}{rgb}{0.93, 0.53, 0.18}

\title{Language Models on a Diet: Cost-Efficient Development of Encoders for Closely-Related Languages via \\Additional Pretraining}

\name{Nikola Ljubešić\textsuperscript{1,2}, Vít Suchomel\textsuperscript{3}, Peter Rupnik\textsuperscript{1}, Taja Kuzman\textsuperscript{1}, Rik van Noord\textsuperscript{4}}

\address{\textsuperscript{1}Jožef Stefan Institute, \textsuperscript{2}University of Ljubljana, \textsuperscript{3}Masaryk University, \textsuperscript{4}University of Groningen \\
         nikola.ljubesic@ijs.si, vit.suchomel@sketchengine.eu,\\peter.rupnik@ijs.si, taja.kuzman@ijs.si, r.i.k.van.noord@rug.nl\\
         }

\abstract{
The world of language models is going through turbulent times, better and ever larger models are coming out at an unprecedented speed. 
However, we argue that, especially for the scientific community, encoder models of up to 1 billion parameters are still very much needed, their primary usage being in enriching large collections of data with metadata necessary for downstream research. We investigate the best way to ensure the existence of such encoder models on the set of very closely related languages -- Croatian, Serbian, Bosnian and Montenegrin, by setting up a diverse benchmark for these languages, and comparing the trained-from-scratch models with the new models constructed via additional pretraining of existing multilingual models. We show that comparable performance to dedicated from-scratch models can be obtained by additionally pretraining available multilingual models even with a limited amount of computation. We also show that neighboring languages, in our case Slovenian, can be included in the additional pretraining with little to no loss in the performance of the final model.
\\ \newline 
}

\begin{document}

\maketitleabstract





\section{Introduction}

The field of natural language processing is in the middle of a paradigm shift due to the emergence of large language models (LLMs) that showcase impressive capabilities across a diverse range of natural language understanding tasks.
While the current front-runners mainly cover English and some other `large' languages \citep{OPT, GPT4, LLAMA}, it is just a matter of time for those models to start performing on a similar (or even higher) level for less-resourced languages. One example is the COPA benchmark for South Slavic languages. This task was just partially solvable by smaller non-English language models~\cite{ljubesic-lauc-2021-bertic}, to which GPT-3.5 Turbo has been catching up significantly even for very under-resourced languages such as Macedonian. What is more, GPT-4 was shown to bring the performance for all South Slavic languages to the level of its performance on the English version of the same benchmark.\footnote{\url{https://github.com/clarinsi/benchich/tree/main/copa}}

With these developments, we are placed today in front of a big dilemma. Should we simply wait for large language models to become more parameter- and data-efficient, thereby encompassing our languages of interest with good-enough performance? Alternatively, is there still room for the up-to-1-billion-parameters models that we are able to pretrain with the limited computing capacity available in most of academia? Our claim is that, besides the pure academic endeavor of researching language modelling techniques, which are very needed activities by themselves, on the application side there is still a need for encoder models of the up-to-1-billion-parameters size, primarily for the enrichment of our research data, mostly large corpora, for downstream research. Examples of such enrichment are genre annotation of tens of millions of documents inside the CLASSLA web corpora of South Slavic languages with the X-GENRE Transformer-based classifier~\cite{kuzman2023automatic}, or annotation of billions of tokens of the ParlaMint corpus of parliamentary proceedings with the latest Transformer-based sentiment models~\cite{mochtak2023parlasent}.

In addition to concerns that large language models might simply require too much computation (or even more problematic, API calls) to enrich millions of documents, there are additional issues with using large language models for data enrichment for scientific purposes. These considerations are twofold. Firstly, the decoder models do not generate limited classification or regression outputs, but free text, which is often hard to map to the pre-defined set of classes intended for downstream data analysis. And secondly, they perform overall great in zero-shot, in-context learning scenarios, but as the length of the instruction, provided in a prompt, is very limited, it is not possible to provide detailed directions on how to separate between less clear cases, as can be achieved via manual annotation of thousands of instances, on which fine-tuned encoder models are based ~\citep{kuzman2023automatic}.

\paragraph{Languages in focus} In this paper, we search for the best path towards creating well-performing encoder language models with less than a billion parameters for medium-sized languages. We perform our search on the example of the South Slavic pluricentric Serbo-Croatian macro-language (code \texttt{hbs} by ISO 639-3, called HBS onward). The HBS macro-language encompasses the following official languages: Bosnian (code \texttt{bs} by ISO 639-1), Croatian (\texttt{hr} by ISO 639-1), Montenegrin (\texttt{cnr} by ISO 639-3) and Serbian (\texttt{sr} by ISO 639-1). We investigate the following options: (1) pretraining the models from scratch, as is the case with the BERTić model~\citep{ljubesic-lauc-2021-bertic}, pretrained on more than 8 billion words of Croatian, Bosnian, Montenegrin and Serbian texts, or the cseBERT model~\cite{ulcar-robnik2020finest}, pretrained on Slovenian, English and Croatian texts, and (2) additionally pretraining existing multilingual models, specializing them for the languages of interest.

\paragraph{Research questions} To explore the second option, we additionally pretrain base-sized and large-sized XLM-RoBERTa (XLM-R) models~\cite{conneau2020unsupervised} with a comparable amount of computation. Furthermore, we compare the model additionally pretrained on HBS data only, as well as a model additionally pretrained on both HBS and Slovenian, a closely-related, but not mutually intelligible South Slavic languages.
The main questions that we want to obtain an answer for are the following:
(1) Is it possible to achieve performance of dedicated models that were trained-from-scratch (BERTić or cseBERT) by additionally pretraining a multilingual model (XLM-R) for a limited number of steps? (2) How do base and large XLM-R models compare in this approach? (3) Is it beneficial not to additionally pretrain for a single language, but include closely related languages into the additional pretraining as well?




\paragraph{Contributions} The contributions of this paper are the following: (1) we expand an existing benchmark~\cite{rupnik-etal-2023-benchic}\footnote{\url{https://github.com/clarinsi/benchich/}} with three additional tasks, one for named entity recognition on four separate datasets, another for sentiment identification on political texts, and a final one on causal commonsense reasoning on two datasets, (2) we build the largest collection of raw HBS text up to this point, measuring 11.5 billion words,\footnote{\url{https://huggingface.co/datasets/classla/xlm-r-bertic-data}} (3) we obtain insights into how base and large multilingual models behave as they get additionally pretrained, comparing the pretraining on a single language group (HBS) and the language group extended with a closely related language (Slovenian), and, finally, (4) we release new models for the HBS languages\footnote{\url{https://huggingface.co/classla/xlm-r-bertic}} as well as for Slovenian and the HBS languages\footnote{\url{https://huggingface.co/classla/xlm-r-bertic}} which achieve comparable or improved performance on the four tasks.



\section{Related Work}

Given the significant impact of BERT~\citep{devlin-etal-2019-bert}, there has been a large push towards similarly effective models for all other languages, especially given the often inferior performance of the multilingual BERT variant for low-resource languages \citep{wu2020all}. Following these findings, researchers started exploring how to cater to low-resource languages. We can see three major approaches: 1) development of monolingual models, 2) development of moderately multilingual models, 3) adapting massively multilingual models to improve their performance on the target language.

\paragraph{Monolingual models}
Monolingual models are pretrained from scratch on texts in one language. Given the relative simplicity of this approach and the initial effectiveness in terms of downstream performance, many successful monolingual language models (LMs) were developed \citep{bertje, camembert, flaubert, estbert, icebert}. While monolingual models often provided the best performance \citep{ulvcar2021evaluation}, in the case of less-resourced languages,  the main limitation of this approach is that there might not be enough available data for pretraining.

\paragraph{Moderately multilingual models}
To mitigate this challenge, development of moderately multilingual models was suggested \citep{ulcar-robnik2020finest}. In this case, the model is pretrained from scratch as well, but on data from multiple closely-related languages.
This approach was used in \citet{ulcar-robnik2020finest}, who developed the CroSloEngual BERT (cseBERT) model which was pretrained on three languages: Croatian and Slovenian, which are closely related, and English. Similarly, the BERTić model \citep{ljubesic-lauc-2021-bertic} 
was pretrained on four languages that are very closely related and mutually intelligible: Bosnian, Croatian, Serbian and Montenegrin. 
This model outperformed cseBERT on downstream tasks in Croatian (except on named entity recognition), as was shown in \citet{ulvcar2021evaluation}, likely because it was trained on significantly more data.
\citet{singh2023too} experimented with bilingual models and showed that they outperform the massively multilingual models even if the two languages that are combined for training are very distant, e.g., Slovenian and Basque.
Additionally, as these models are multilingual, they can be used in cross-language learning scenarios between the included languages \citep{ulcar-robnik2020finest}.
Furthermore, this is a more cost-efficient approach, as it accommodates multiple low-resource languages with the cost of pretraining a single model.

\paragraph{Adaptation} However, both these approaches demand pretraining models from scratch, which is very computationally expensive. To mitigate this, one can benefit from existing massively multilingual pretrained models and simply adapt them to
the target low-resource language.
There are two main approaches for adaptation of massively multilingual models to specific languages: 1) language-adaptive pretraining 
and 2) adapters \citep{pfeiffer2020mad}. In the case of language-adaptive pretraining the massively multilingual model is additionally pretrained with the masked language modelling (MLM) objective on data in the target language. 
This method was repeatedly shown to provide better results than the base massively multilingual model on monolingual tasks \citep{wang2020extending,chau2020parsing,icebert}. An alternative method 
is adapting massively multilingual models to specific languages by learning modular language-specific representations via adapters \citep{pfeiffer2020mad,pfeiffer2021adapterfusion}. 
\citet{ebrahimi2021adapt} compared the methods of extending XLM-RoBERTa to low-resource languages on multiple NLP tasks in a cross-language zero-shot scenario. They showed that additional pretraining provides the best results, while considering it also to be the simplest method to apply. 
Moreover, additionally pretraining requires much less pretraining than pretraining a model from scratch, and is thus more cost-efficient. Consequently, we have decided to employ this method in the development of language models for the HBS macro-language and Slovenian language. An additional motivation for this choice is the fact that this particular approach has not yet been explored in the context of South Slavic languages.

\section{Additional Pretraining}

\subsection{Data}

In this section, we describe the data used for additional pretraining of the XLM-RoBERTa models. We separately describe the HBS and the Slovenian data collection. These two collections jointly consist of more than 19 billion words of running text.
All the data inside each language group are heavily near-deduplicated by using Onion\footnote{\url{https://corpus.tools/wiki/Onion}}~\citep{pomikalek2011removing} with 5-tuples of words, a 90\% duplicate threshold and smoothing disabled. The tool operates on the paragraph level, provided that the paragraphs are available (originally separated either as HTML block elements or empty lines), otherwise on the document level.

\paragraph{HBS} For the HBS collection of languages, we compiled, to the best of our knowledge, the largest collection of HBS texts up to this date, consisting of 11.5 billion words of running text. The collection consists, in order of near-deduplication\footnote{The order of near-deduplication is important because it works on the "first-come-only-retained" principle, only the first paragraph of mutually similar text being retained, all later occurring paragraphs being removed from the collection.}, of the recent MaCoCu crawl of the Croatian~\citep{macocu-hr}, Bosnian~\citep{macocu-bs}, Montenegrin~\citep{macocu-cnr} and Serbian web~\citep{macocu-sr}; the text collection on which the BERTić model~\cite{ljubesic-lauc-2021-bertic} was pretrained -- including the hrWaC, slWaC, srWaC, and bsWaC web corpora~\cite{ljubevsic2011hrwac,ljubesic-klubicka-2014-bs}, the CC100 collection~\cite{conneau2020unsupervised}, and the Riznica corpus~\citep{riznica} --; a collection of on-line newspapers donated for the purpose of training the presented models; and the mC4 collection~\citep{xue-etal-2021-mt5}. The size of each part of the HBS pretraining data is given in Table~\ref{tab:hbs_data}. One should note that while the BERTić data collection was originally 8.39 billion words large, its size has shrunk to 3.82 billion words due to the harsh near-deduplication especially with the recent MaCoCu crawls, which certainly contain older web data as well. A similar phenomenon can be observed for the mC4 dataset, which was originally 1.74 billion words in size, shrinking down to 800 million words only.

\begin{table}[ht]
\begin{center}
  \begin{tabular}{lr}
    \toprule
   \textbf{Dataset} & \textbf{Number of words} \\
    \midrule
    MaCoCu HBS & 5,490,335,790 \\
    BERTić data & 3,815,720,806 \\
    Online newspaper & 1,433,110,363 \\
    mC4 &  799,773,550 \\
    \midrule
    Total & 11,538,940,509 \\
  \bottomrule
\end{tabular}
  \caption{Overview of the pretraining data for the HBS language group.}
  \label{tab:hbs_data}
\end{center}
\end{table}

\paragraph{Slovenian} For Slovenian we primarily, again in the order of near-deduplication, relied on the recent MaCoCu crawl of the Slovenian web~\citelanguageresource{macocu-sl}, but also included the very large MetaFida corpora collection~\citep{metafida} (including, but not limited to the reference GigaFida corpus~\citep{krek-etal-2020-gigafida} and the KAS corpus of academic writing~\cite{erjavec2021kas}), as well as the mC4 dataset~\citep{xue-etal-2021-mt5} and the CC100 dataset~\citep{conneau2020unsupervised}. An overview is shown in Table~\ref{tab:sl_data}.

\begin{table}[ht]
\begin{center}
  \begin{tabular}{lr}
    \toprule
    \textbf{Dataset} & \textbf{Number of words} \\
    \midrule
    MaCoCu Slovenian & 1,907,662,185 \\
    MetaFida & 3,257,795,640 \\
    mC4 & 2,263,513,217 \\
    CC100 & 195,989,576 \\
    \midrule
    Total & 7,624,960,618 \\
  \bottomrule
\end{tabular}
  \caption{Overview of the pretraining data for the Slovenian language.}
  \label{tab:sl_data}
\end{center}
\end{table}

\subsection{Methodology}

We perform additional pretraining of the massively multilingual XLM-RoBERTa (XLM-R)~\citep{conneau2020unsupervised} model in base size (XLM-R-base) and large size (XLM-R-large). The base-sized model we only additionally pretrain on the HBS data collection. Henceforth, this model is referred to as XLM-R-base-BERTić, or XB-BERTić for brevity.  The large model, which is pretrained on the HBS data collection, is denoted as XLM-R-large-BERTić, or XL-BERTić. Additionally, the model pretrained on the merged HBS and Slovenian data collection is named XLM-R-large-SloBERTić, or XL-SloBERTić. 
We perform additional pretraining on the Google Cloud infrastructure, using a single TPUv3 for each pretraining with a batch size of 1,024. We run each pretraining process with a comparable amount of computation. For the base model, we perform 96k steps overall, while for large models we perform 48k steps. We organize each pretraining into 8 rounds and report the results at the end of each round. A description of models with additional pretraining hyperparameters is shown in Table~\ref{tab:additional_pretraining}.


\begin{table}[]
    \centering
   \setlength{\tabcolsep}{3.5pt}
    \resizebox{\columnwidth}{!}{
    \begin{tabular}{lllll}
    \toprule
    \bf Name & \bf Data & \bf Steps & \bf Warmup & \bf LR \\ 
    \midrule
         XB-BERTić &  HBS & 96k & 5k & 1e-04 \\
         XL-BERTić &  HBS & 48k & 2.5k & 1e-04 \\
         XL-SloBERTić & HBS + SL & 48k & 2.5k & 1e-04 \\
         \bottomrule
    \end{tabular}
    }
   \caption{Information on the pretraining hyperparameters and data for the newly introduced models. XB-BERTić is the XLM-R-base model additionally pretrained on HBS data only. XL-BERTić is the XLM-R-large model additionally pretrained on HBS data only. XL-SloBERTić is XML-R-large model additionally pretrained on HBS and Slovenian data.}
    \label{tab:additional_pretraining}
\end{table}

\section{Evaluation}

We evaluate the models on three diverse tasks. We use named entity recognition as a token classification task over two Croatian and two Serbian datasets. Next, we evaluate the models on a sequence regression task in form of a parliamentary sentiment prediction task. Lastly, we evaluate on a sequence pair classification task via the choice of plausible alternatives (COPA) dataset translations into Croatian and Serbian. We describe the three tasks in detail below.

\subsection{Datasets}

\paragraph{Named Entity Recognition}

We evaluate the performance of the models on the task of  named entity recognition on two languages -- Croatian and Serbian. Our benchmark consists of two datasets per language: one for the standard language, another for the non-standard language. Specifically, the following datasets are used:

\begin{itemize}
    \item Croatian linguistic training corpus hr500k 2.0 \citep{NER-s-HR}
    \item Croatian Twitter training corpus ReLDI-NormTagNER-hr 3.0 \citep{NER-ns-HR}
    \item Serbian linguistic training corpus SETimes.SR 2.0 \citep{NER-s-SR}
    \item Serbian Twitter training corpus ReLDI-NormTagNER-sr 3.0 \citep{NER-ns-SR}
\end{itemize}

\begin{table}
\centering
\begin{tabular}{lr}
\toprule
\textbf{Dataset }            & \textbf{Number of tokens} \\
\midrule
hr500k              & 499,635      \\
ReLDI-NormTagNER-hr & 89,855       \\
ReLDI-NormTagNER-sr & 97,673       \\
SETimes.SR          & 92,271      \\
\bottomrule
\end{tabular}
\caption{\label{ner-datasets} Sizes of datasets (in tokens), used in the named entity recognition experiments.
}
\end{table}

We use the train, development and test set splits as they are split in the original datasets.

\paragraph{Sentiment Identification}
For experiments on sentiment, we use the ParlaSent dataset~\citep{parlasent2023}, a dataset of sentences from parliamentary proceedings, manually annotated for sentiment. Specifically, we use the HBS train and test subsets, each of them containing 2,600  sentences annotated with an ordinal 0 (negative) to 5 (positive) schema.


\paragraph{Commonsense Reasoning}

The Choice of Plausible Alternatives (COPA, \citealp{copa}) is a task in which a model has to choose between two plausible continuations of text, given a premise sentence, and return the more plausible one. This task is part of the SuperGLUE English benchmark \citep{wang2019superglue} and has human translations available for Croatian~\citep{copahr} and Serbian~\citep{copasr}. We use the standard split of 400 training, 100 development and 500 test instances.

\subsection{Evaluation Methodology}

\paragraph{Baseline Models}

We compare our newly introduced models to four baseline models: two moderately multilingual models, BERTić~\citep{ljubesic-lauc-2021-bertic} and cseBERT~\citep{ulcar-robnik2020finest}, and the massively multilingual XLM-RoBERTa (XLM-R)~\citep{conneau2020unsupervised} model in base and large size. The BERTić model was pretrained on 8.4 billion words in mostly Croatian, but also very closely related, mutually intelligible languages of Bosnian, Serbian and Montenegrin~\cite{ljubesic-lauc-2021-bertic}. The cseBERT model was pretrained on 5.9 billion tokens, of which 31\% were in Croatian, 23\% in Slovenian, and the rest in English.  The massively multilingual XLM-R model was pretrained on the CommonCrawl multilingual data~\citep{conneau2020unsupervised}, which consists of 167 billion tokens in 100 languages. In terms of the size, the BERTić and cseBERT models are comparable to the base-sized XLM-R with 12 hidden layers and 768 hidden states, whereas the large-sized XLM-R is approximately three times larger in terms of the number of parameters, and consists of 24 hidden layers and 1,024 hidden states.

\paragraph{Hyperparameter Search}


For all tasks, we perform hyperparameter searches for the BERTić model, the cseBERT model, the base-sized XLM-R model and the large-sized XLM-R model. For the newly introduced models, the best settings of XLM-R-base are used for XB-BERTić, while the settings of XLM-R-large were used for XL-BERTić and XL-SloBERTić.
In both named entity recognition and sentiment identification, we optimize only the learning rate and the number of epochs. The hyperparameter search is performed by evaluating on the development data. For named entity recognition, optimal hyperparameters depend on the NER dataset. We perform a separate hyperparameter search for the Croatian standard dataset and for the Serbian standard dataset because of the difference in size, while we perform a joint hyperparameter search for the two non-standard datasets due to their very similar size and diversity. For sentiment identification, we perform a hyperparameter search on a subset of the training dataset which is marked as validation data, as defined in the ParlaSent dataset~\citep{mochtak2023parlasent}. For COPA, we perform a hyperparameter search over learning rate and batch size.  During fine-tuning, we always train for 15 epochs. Detailed hyperparameter settings are shown in Section \ref{app:hyperparams} in the Appendix. 

\paragraph{Evaluation Setup}

For named entity recognition, we train and test each model three times and report aggregated results in the macro F1 score. For sentiment, we perform five runs, and report average $R^2$ scores. For COPA, we average over 10 runs and report the accuracy score. 

\section{Results}

In this section, we present the results of the evaluation of the newly trained models, compared to the existing models that were trained from scratch, namely BERTić, cseBERT, XLM-R-base and XLM-R-large. We consider these four models as the baseline models. Additionally, to provide insights into the efficiency of pretraining, we do not only evaluate the final pretrained model -- we evaluate models, created in 8 rounds of additional pretraining, where base models are updated for 12k steps per round and large models 6k steps, each round corresponding to an identical amount of computation regardless of model size. We evaluate the models on three tasks: the token classification task of named entity recognition, the sequence regression task of sentiment analysis, and the sequence pair classification task in form of the commonsense reasoning benchmark COPA.

\subsection{Named Entity Recognition}

Given that the named entity recognition task consists of four datasets, here we present a summarized version of the results in form of average results on the two standard and the two non-standard datasets. The full results are available in Section \ref{subsubsection:NER} in the Appendix.

\begin{figure*}
    \begin{subfigure}{0.45\textwidth}
        \includegraphics[width=\textwidth]{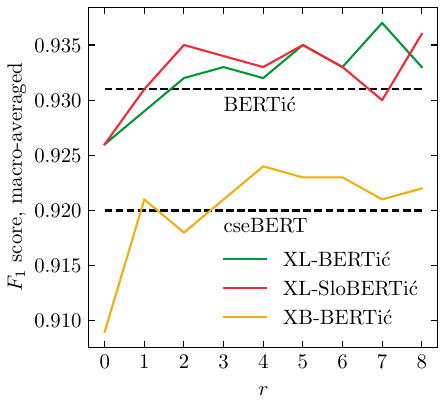}
        \caption{Standard NER\vspace{0.5cm}}
        \label{fig:nerstd}
    \end{subfigure}
    \hfill
    \begin{subfigure}{0.45\textwidth}
        \includegraphics[width=\textwidth]{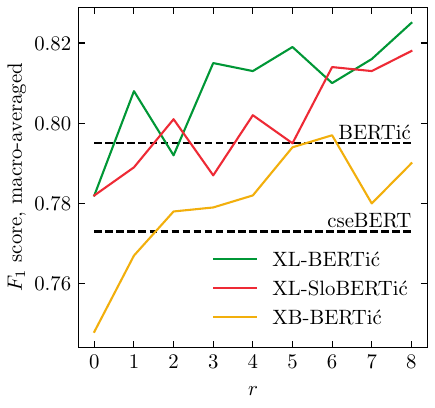}
        \caption{Non-standard NER\vspace{0.5cm}}
        \label{fig:nernstd}
    \end{subfigure}
    \begin{subfigure}{0.45\textwidth}
        \includegraphics[width=\textwidth]{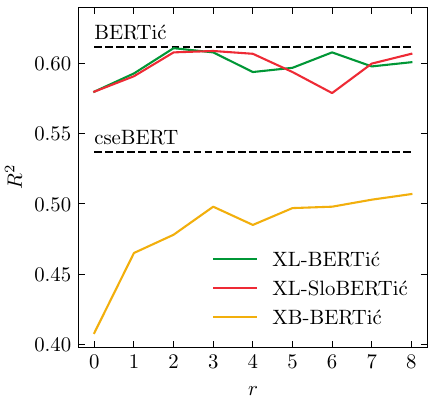}
        \caption{Sentiment regression}
        \label{fig:sentiment}
    \end{subfigure}
    \hfill
    \begin{subfigure}{0.45\textwidth}
        \includegraphics[width=\textwidth]{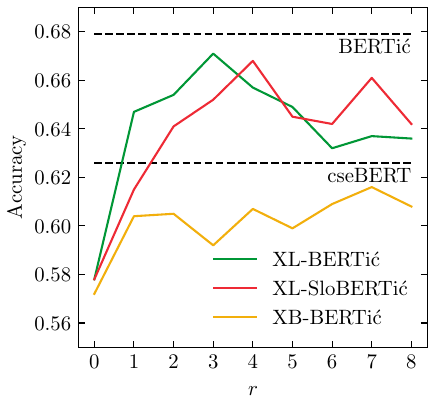}
        \caption{Causal commonsense reasoning}
        \label{fig:copa}
    \end{subfigure}
    \caption{ Performance of models on different tasks in relation to the round of additional pretraining.
    $r=0$ is referring to round 0, before any additional pretraining, and thus represents the performance of the XLM-RoBERTa-base and XLM-RoBERTa-large models. Subsequent 8 datapoints represent stages of additional pretraining. One round equals 12k steps for the base model (XB-BERTić), and 6k steps for large models (XL-BERTić and XL-SloBERTić),  in this way identical amount of computation per round was assured regardless of model size. The performance of cseBERT and BERTić is depicted with a black dashed line.}
\end{figure*}


\paragraph{Standard datasets} Figure~\ref{fig:nerstd} presents the performance of all the compared models on the two standard named entity recognition datasets. From the baseline models, BERTić performs the best, with a minor difference to cseBERT. XLM-R-large performs between the two models, while the  XLM-R-base model underperforms. Once the XLM-R models are additionally pretrained, their performance significantly improves, with the biggest improvements being achieved in the first few rounds of additional pretraining. When we compare the BERTić and the SloBERTić versions of the updated XLM-R-large models to these baselines, we do not see any difference in performance. Full results are published in  Section \ref{subsubsection:NER} in the Appendix.

\paragraph{Non-standard datasets} When the models are evaluated on the two non-standard datasets, results of which are presented in Figure~\ref{fig:nernstd}, the picture is somewhat similar to the results on the standard datasets. Among the baseline models, BERTić performs best, with XLM-R-large positioned between BERTić and cseBERT. XLM-R-base again shows significantly lower results. Updating the XLM-R models shows that the models' performances improve most in the first rounds of additional pretraining, with the difference to the standard data that the models' improvement does not completely flatten out, but raises slightly through all of the 8 rounds of additional pretraining. An early hypothesis for this behavior is that non-standard named entity recognition is a harder task and observing more data during additional pretraining has a slight positive effect, one that cannot be observed when performing named entity recognition over standard data.
Full results are published in Section \ref{subsubsection:NER} in the Appendix.

\paragraph{Overall NER results} Overall, on both the standard and the non-standard dataset collections, the additionally updated XLM-R-large improves slightly over the best-performing out of all the baseline models, which is the BERTić model. This improvement is more pronounced on the non-standard datasets.

\subsection{Sentiment Identification}

\paragraph{Baselines} Secondly, we evaluate the models on sentiment identification on parliamentary proceedings. In Figure~\ref{fig:sentiment}, we present our results in a comparable manner to the named entity recognition results. The results of the baseline models are comparable to the NER results. That is, BERTić achieves the best results, XLM-R-large falls somewhere between BERTić and cseBERT, while the base-sized XLM-R performs the worst.

\paragraph{Additional pretraining} Additional pretraining shows a very similar behavior to the NER results on the standard language datasets. Namely, XLM-R-large models improve their results mostly during the first few rounds of additional pretraining, the improvements being leveled out further. However, a clear difference is that the XLM-R-base model this time achieves improvements throughout all the 8 rounds of additional pretraining. Regarding the difference in performance between the XL-BERTić and the XL-SloBERTić model, the results are comparable to those in the named entity recognition task, with almost no negative impact if significant part of the pretraining was performed on a closely related language. For the overall best results, the updated XLM-R-large model never surpasses, but  arrives close to the result of the best-performing BERTić model.
Full results are published in Section \ref{subsubsection:sentiment} in the Appendix.

\subsection{Commonsense Reasoning}
\paragraph{Baselines} In this subsection, we present the results over our two commonsense reasoning datasets, COPA-HR and COPA-SR in Figure~\ref{fig:copa}. If we compare the baseline models, we can see that while BERTić still performs the best, cseBERT now positions itself as the second-best system, in contrast to the results in the two previous tasks. Here, XLM-R-large shows significantly lower performance than BERTić and cseBERT. This is in agreement with previous results, showing that multilingual models for smaller languages, such as Croatian and Serbian, do not perform well on the COPA task~\cite{ljubesic-lauc-2021-bertic}. Interestingly enough, there is not a big difference in performance of the large-sized and the base-sized XLM-R model.

\paragraph{Additional pretraining} Once the XLM-R models undergo additional pretraining, their performance exhibits a significant improvement during the initial rounds of updates. However, an unexpected phenomenon occurs thereafter, as the models begin to exhibit a decline in performance compared to the early rounds of updates. Although the performance does not regress to the level observed prior to the additional pretraining, the decrease in performance cannot be disregarded. In the subsequent subsection, we discuss this phenomenon further, together with a concise summary of the results obtained across all three tasks. Full results are published in Section \ref{subsubsection:copa} in the Appendix.


\subsection{Discussion}

\paragraph{Baselines} Summarizing the performance of baseline models, we have a clear overall winner -- the BERTić model, which obtains the best result on all tasks and datasets. This follows the previous results of ~\citet{ljubesic-lauc-2021-bertic}, but not those of ~\citet{ulvcar2021evaluation}, the latter potentially not having invested enough in hyperparameter search 
for ELECTRA models. cseBERT does come second in one task -- commonsense reasoning, while in the two remaining tasks XLM-R-large shows to be more potent. The base-sized XLM-R is regularly the worst performing model.

\paragraph{Performance over time} A very interesting trend can be observed when summarizing the results of the additionally pretrained models. What is common to all results, regardless of the task, is that the big improvement in performance comes after just a few rounds of updates.  Once pretraining is continued, the behavior of the models is different depending on the task. When the models are fine-tuned for named entity recognition, which can be regarded as the shallowest task, there is visible improvement throughout the whole additional pretraining process. On the sentiment identification task, such continuous improvements cannot be observed and the performance curve flattens out after a few rounds of additional pretraining. Most interestingly, on the causal commonsense reasoning, which is the most complex task of the three, prolonged training starts to negatively impact the models' performance. Our early hypothesis for this very interesting phenomenon is the following: additional pretraining of XLM-R models just with a single language (group), if performed for long enough, starts to break the multilingual fabric of the model. Considering that the majority of the collective knowledge has been acquired from the ``large'' languages, which are most prominent in the pretraining data of the XLM-R models, deviating from this shared representation, by pretraining on less prominent languages, results in the loss of crucial profound knowledge required for tasks like commonsense reasoning. The adverse impact is not observable in less complex tasks such as named entity recognition, where the use of shared multilingual knowledge is relatively low, and the additional pretraining compensates for the loss incurred by diverging from the multilingual representations. 

\paragraph{Adding related languages} Furthermore, in the evaluation, we also compare the performance of the models that were additionally pretrained on the HBS language group with the models where we included also Slovenian in the pretraining data. While Slovenian is closely related to HBS, it is not mutually intelligible with the languages in the HBS language group. 
The results show that there is no negative impact to the model's performance if closely related languages are also included in the training data, and thus indicate that the cost-efficiency of developing encoder models for less resourced languages can be yet further improved by additionally pretraining on multiple related languages and providing for them all at once.

\section{Conclusion}

\paragraph{Summary} This paper investigates how dedicated monolingual or moderately multilingual encoder models that were pretrained from scratch compare to additionally pretraining massively multilingual encoder models of size up to 1 billion parameters on the example of the HBS language group, comprising the Bosnian, Croatian, Montenegrin and Serbian official languages. The existing and newly introduced models for HBS are evaluated on a benchmark that comprises a token classification task (named entity recognition), a sequence regression task (sentiment analysis) and a sequence pair classification task (causal commonsense reasoning). The benchmark is available at \url{https://github.com/clarinsi/benchich/} and we invite the research community to add additional models to this benchmark. Our results show that by additionally pretraining the XLM-R-large model performance on the languages of interest increases significantly on all tasks. However, beyond a certain threshold of additional pretraining, the performance gains begin to level off. In fact, for the task of commonsense reasoning, the performance even decreases. Our hypothesis is that the loss in performance through additional pretraining can be attributed to the potential disruption of the multilingual aspect of the original model, where the majority of the language understanding capacity is encoded. 

\paragraph{Research questions} For our research questions stated in the introduction, we propose the following answers: (1) it is possible to achieve a comparable or even  better performance to the language-specific models trained from scratch if one  additionally pretrains large multilingual models on the language of interest, (2) large multilingual models regularly perform better than the base-sized models, and (3) no drop in performance can be observed if a significant part of the additional pretraining data consists of a closely related language.

\paragraph{Model and data releases} We have decided to publish the two new, additionally pretrained models via HuggingFace -- the XL-BERTić model \url{https://huggingface.co/classla/xlm-r-bertic} and the XL-SloBERTić model \url{https://huggingface.co/classla/xlm-r-slobertic}, both after 48 thousand steps of additional pretraining where most stable results are obtained on all three benchmarking tasks. The reasons for publishing these models are the following: (1) these models perform slightly worse on two, but improve on one task (on both subtasks) to the overall winner of our experiments, the BERTić model, (2) while the BERTić model still performs slightly better on two tasks, we expect for  the XL-SloBERTić model to cover both HBS and Slovenian similarly well, including also cross-lingual learning, both of which still have to be confirmed in upcoming experiments, but are sensible expectations, (3) the new XL-BERTić and XL-SloBERTić models were pretrained on newer data, spanning into 2023, while the BERTić model was pretrained on data spanning until 2019, and (4) the XL-BERTić and XL-SloBERTić models are three times the size of the BERTić model, a feature that might be useful in learning some tasks. We also release the 11.5 billion words of HBS data the models were additionally pre-trained on as a HuggingFace Dataset: \url{https://huggingface.co/datasets/classla/xlm-r-bertic-data}.

\paragraph{Main takeaway} Given the observed results during our experiments, our recommendation for future activities in terms of developing encoder models of up to 1 billion parameters for less-resourced languages 
is for researchers to take advantage of the existing massively multilingual models and specialize them for the language of interest via additional pretraining. 
During additional pretraining, it is important that the performance of the model is continuously analysed via evaluation on relevant tasks. This is important as our findings suggest that after a specific amount of additional pretraining, performance could start to deteriorate due to the loss of deeper language understanding that is provided by the multilingual aspect of the model.
This ``drifting away'' phenomenon might be countered by adding some data of large languages to the dataset used for additional pretraining, but this assumption has to be assessed in future research.

\paragraph{On the notion of under-resourcedness} We have to note that, while the languages in question are less resourced than most of the European and large world languages, they are still not close to under-resourced on the global scale. All the languages in question have been present during pretraining of the XLM-R models, and we performed experiments with additionally pretraining them on multiple billions of words, most of the world languages cannot come close to. However, we are of the position that there is a significant number of  languages that can be helped with the insights provided in this paper.

\section{Acknowledgments}

This work has received funding from the European Union's Connecting Europe Facility 2014-2020 - CEF Telecom, under Grant Agreement No. INEA/CEF/ICT/A2020/2278341. This communication reflects only the author's view. The Agency is not responsible for any use that may be made of the information it contains.

This work was also funded by the Slovenian Research Agency within the basic research project MEZZANINE ``Basic Research for the Development of Spoken Language Resources and Speech Technologies for the Slovenian Language'' (J7-4642) and the research programme ``Language resources and technologies for Slovene'' (P6-0411).

This research was supported with Cloud TPUs from Google's TPU Research Cloud (TRC).



\section{Bibliographical References}\label{reference}

\bibliographystyle{lrec_natbib}
\bibliography{bibliography}

\section{Language Resource References}
\label{lr:ref}
\bibliographystylelanguageresource{lrec_natbib}
\bibliographylanguageresource{languageresource}



\appendix

\section{Appendix}

\subsection{Hyperparameters}
\label{app:hyperparams}

We use the following hyperparameters for fine-tuning the models for the evaluation tasks:
\begin{itemize}
    \item \textbf{Named Entity Recognition}: we use the learning rate of \texttt{4e-05}, the train batch size of \texttt{32} and the maximum sequence length of \texttt{256}. The hyperparameter search showed that optimum number of epochs depends on the size and difficulty level of the named entity dataset. Thus, different numbers of epochs are used depending on the dataset, as shown in Table \ref{table-ner-epochs}.
    \item \textbf{Sentiment Identification}: the hyperparameter search showed that the optimal epoch number for all models is \texttt{15}. We use the train batch size of \texttt{32} and the maximum sequence length of \texttt{256}. In contrast to the named entity recognition task, the optimum learning rate was shown to depend on the model. Namely, we use \texttt{4e-05} for cseBERT and BERTić, and \texttt{8e-06} for the base- and large-sized XLM-RoBERTa models and all additionally pretrained models.
    \item \textbf{Commonsense Reasoning}: we performed a hyperparameter search over batch size and learning rate over the baseline models per language. We actually found uniform results. The best settings were a batch size of \texttt{8} and learning rate of \texttt{1e-05} for a training time of \texttt{15} epochs across all models. Note that when averaging over 10 runs, we ignore failed runs, i.e. runs for which the training loss never decreases. We noticed that this occurred more frequently for the models that were trained for longer.
\end{itemize}

\begin{table}[!ht]
\centering
\begin{tabular}{llll}
\toprule
\bf Model                                  & \bf HR-s & \bf Non-s & \bf SR-s  \\
\midrule
XLM-R-base                  & 5              & 8                            & 6              \\
XLM-R-large & 7              & 11                           & 13             \\
BERTić                                & 9              & 10                           & 10             \\
CSEbert                                & 4              & 7                            & 9            \\
\bottomrule
\end{tabular}
\caption{\label{table-ner-epochs} Epoch number used for fine-tuning the models on different named entity recognition datasets: standard Croatian (HR-s), standard Serbian (SR-s), and non-standard Croatian and Serbian datasets (Non-s). All XB-BERTić models use the same epoch number as the base-size XLM-RoBERTa model (XLM-R-base), and the other pretrained models (XL-BERTić and XL-SloBERTić) use the same epoch number as the large-sized XLM-RoBERTa model (XLM-R-large).
}
\end{table}

\subsection{Full Results}

In the following subsections, we provide more details on the results for all the three tasks, that is, named entity recognition, sentiment identification and commonsense reasoning.

\subsubsection{Named Entity Recognition}
\label{subsubsection:NER}

In this section, we show the results of the evaluation of the models on the named entity recognition task on each of the four evaluated datasets. More precisely, Table \ref{ner-hr500} shows the results on the standard Croatian dataset, Table~\ref{ner-reldi-hr} on non-standard Croatian dataset, Table~\ref{ner-set-sr} on standard Serbian dataset, and Table~\ref{ner-reldi-sr} on non-standard Serbian dataset. We train and test each model three times and report aggregated results, using the macro F1 score.

\begin{table*}[!ht]
\centering
\begin{tabular}{l|l|lllll}
\hline
\bf base & \bf large & \bf cseBERT     & \bf BERTić      & \bf XB-BERTić & \bf XL-BERTić & \bf XL-SloBERTić \\
\hline
0    & 0     & 0.918±0.002 & 0.925±0.003 & 0.903±0.001 & 0.919±0.005 & 0.919±0.005    \\
12   & 6     &             &             & 0.915±0.001 & 0.917±0.005 & 0.920±0.007    \\
24   & 12    &             &             & 0.911±0.004 & 0.923±0.004 & 0.926±0.001    \\
36   & 18    &             &             & 0.912±0.004 & 0.918±0.005 & 0.922±0.005    \\
48   & 24    &             &             & 0.916±0.007 & 0.921±0.002 & 0.926±0.001    \\
60   & 30    &             &             & 0.916±0.001 & 0.929±0.005 & 0.925±0.004    \\
72   & 36    &             &             & 0.916±0.001 & 0.929±0.002 & 0.925±0.004    \\
84   & 42    &             &             & 0.918±0.004 & 0.926±0.003 & 0.927±0.003    \\
96   & 48    &             &             & 0.917±0.002 & 0.927±0.001 & 0.923±0.006 \\
\hline
\end{tabular}
\caption{\label{ner-hr500} Comparison of the models on the NER task on the standard Croatian dataset (hr500k) in terms of macro F1 score, averaged over 3 runs. `base' and `large' correspond to the number of steps performed to additionally pretrain base- or large-sized models, each row therefore requiring equal amount of time on a TPU. Step 0 in the columns for XB-BERTić, XL-BERTić and XL-SloBERTić represents the performance of the models prior to pretraining, i.e., the performance of the XLM-RoBERTa-base and XLM-RoBERTa-large.
}
\end{table*}

\begin{table*}[!ht]
\centering
\begin{tabular}{l|l|lllll}
\hline
\bf base & \bf large & \bf cseBERT     & \bf BERTić      & \bf XB-BERTić & \bf XL-BERTić & \bf XL-SloBERTić \\
\hline
0    & 0     & 0.794±0.006 & 0.792±0.016 & 0.763±0.016 & 0.791±0.014 & 0.791±0.014    \\
12   & 6     &             &             & 0.768±0.010                         & 0.810±0.021 & 0.789±0.034    \\
24   & 12    &             &             & 0.770±0.018                         & 0.810±0.003 & 0.805±0.034    \\
36   & 18    &             &             & 0.790±0.024                         & 0.818±0.015 & 0.802±0.021    \\
48   & 24    &             &             & 0.791±0.015                         & 0.810±0.027 & 0.779±0.024    \\
60   & 30    &             &             & 0.786±0.015                         & 0.803±0.013 & 0.802±0.017    \\
72   & 36    &             &             & 0.806±0.005                         & 0.814±0.005 & 0.820±0.003    \\
84   & 42    &             &             & 0.782±0.016                         & 0.797±0.015 & 0.810±0.008    \\
96   & 48    &             &             & 0.792±0.018                         & 0.809±0.032 & 0.812±0.012   \\
\hline
\end{tabular}
\caption{\label{ner-reldi-hr} Comparison of the models on the NER task on the non-standard Croatian dataset (ReLDI-hr) in terms of macro F1 score, averaged over 3 runs. `base' and `large' correspond to the number of steps performed to additionally pretrain base- or large-sized models, each row therefore requiring equal amount of time on a TPU. Step 0 in the columns for XB-BERTić, XL-BERTić and XL-SloBERTić represents the performance of the models prior to pretraining, i.e., the performance of the XLM-RoBERTa-base and XLM-RoBERTa-large.
}
\end{table*}

\begin{table*}[!ht]
\centering
\begin{tabular}{l|l|lllll}
\hline
\bf base & \bf large & \bf cseBERT     & \bf BERTić      & \bf XB-BERTić & \bf XL-BERTić & \bf XL-SloBERTić \\
\hline
0    & 0     & 0.922±0.002 & 0.936±0.004 & 0.914±0.004 & 0.933±0.005 & 0.933±0.005 \\
12   & 6     &             &             & 0.926±0.005                         & 0.942±0.003                         & 0.941±0.010                         \\
24   & 12    &             &             & 0.925±0.006                         & 0.941±0.004                         & 0.944±0.003                         \\
36   & 18    &             &             & 0.930±0.001                         & 0.947±0.005                         & 0.946±0.005                         \\
48   & 24    &             &             & 0.932±0.001                         & 0.944±0.001                         & 0.941±0.005                         \\
60   & 30    &             &             & 0.930±0.003                         & 0.942±0.004                         & 0.945±0.006                         \\
72   & 36    &             &             & 0.929±0.006                         & 0.938±0.003                         & 0.941±0.010                         \\
84   & 42    &             &             & 0.924±0.004                         & 0.948±0.008                         & 0.932±0.008                         \\
96   & 48    &             &             & 0.927±0.004                         & 0.940±0.003                         & 0.949±0.003          \\     \hline
\end{tabular}
\caption{\label{ner-set-sr} Comparison of the models on the NER task on the standard Serbian dataset (SETimes.SR) in terms of macro F1 score, averaged over 3 runs. `base' and `large' correspond to the number of steps performed to additionally pretrain base- or large-sized models, each row therefore requiring equal amount of time on a TPU. Step 0 in the columns for XB-BERTić, XL-BERTić and XL-SloBERTić represents the performance of the models prior to pretraining, i.e., the performance of the XLM-RoBERTa-base and XLM-RoBERTa-large.
}
\end{table*}

\begin{table*}[!ht]
\centering
\begin{tabular}{l|l|lllll}
\hline
\bf base & \bf large & \bf cseBERT     & \bf BERTić      & \bf XB-BERTić & \bf XL-BERTić & \bf XL-SloBERTić \\
\hline
0    & 0     & 0.751±0.012 & 0.798±0.033 & 0.734±0.024 & 0.774±0.013 & 0.774±0.013    \\
12   & 6     &             &             & 0.765±0.005 & 0.806±0.006 & 0.790±0.031    \\
24   & 12    &             &             & 0.786±0.007 & 0.775±0.024 & 0.797±0.014    \\
36   & 18    &             &             & 0.768±0.024 & 0.812±0.010 & 0.772±0.021    \\
48   & 24    &             &             & 0.772±0.006 & 0.816±0.026 & 0.825±0.016    \\
60   & 30    &             &             & 0.802±0.002 & 0.834±0.026 & 0.788±0.021    \\
72   & 36    &             &             & 0.787±0.018 & 0.805±0.064 & 0.809±0.010    \\
84   & 42    &             &             & 0.779±0.005 & 0.834±0.018 & 0.816±0.030    \\
96   & 48    &             &             & 0.788±0.009 & 0.841±0.013 & 0.824±0.006 \\
\hline
\end{tabular}
\caption{\label{ner-reldi-sr} Comparison of the models on the NER task on the non-standard Serbian dataset (ReLDI-sr) in terms of macro F1 score, averaged over 3 runs. `base' and `large' correspond to the number of steps performed to additionally pretrain base- or large-sized models, each row therefore requiring equal amount of time on a TPU. Step 0 in the columns for XB-BERTić, XL-BERTić and XL-SloBERTić represents the performance of the models prior to pretraining, i.e., the performance of the XLM-RoBERTa-base and XLM-RoBERTa-large.
}
\end{table*}

\subsubsection{Sentiment Identification}
\label{subsubsection:sentiment}
Table \ref{parlasent} shows the results of evaluation of the models on the task of sentiment identification on parliamentary data. We train and test each model five times and report average $R^2$ scores.

\begin{table*}[!ht]
\centering
\begin{tabular}{r|r|rrrrr}
\hline
\bf base & \bf large & \bf cseBERT     & \bf BERTić      & \bf XB-BERTić & \bf XL-BERTić & \bf XL-SloBERTić \\
\hline
    0 & 0 & 0.537±0.006 & 0.612±0.005 & 0.408±0.007 & 0.580±0.014 & 0.580±0.014 \\
    12	& 6 & & & 0.465±0.009 & 0.593±0.009 & 0.591±0.010 \\
    24 & 12 & & & 0.478±0.006 & 0.611±0.004 & 0.608±0.006 \\
    36 & 18 & & & 0.498±0.011 & 0.608±0.009 & 0.609±0.007\\
    48 & 24 & & & 0.485±0.010 & 0.594±0.006 & 0.607±0.008 \\
    60 & 30 & & & 0.497±0.003 & 0.597±0.009 & 0.594±0.005 \\
    72 & 36 & & & 0.498±0.009 & 0.608±0.012 & 0.579±0.055 \\
    84 & 42 & & & 0.503±0.003 & 0.598±0.008 & 0.600±0.006 \\
    96 & 48 & & & 0.507±0.008 & 0.601±0.007 & 0.607±0.007 \\
\hline
\end{tabular}
\caption{Comparison of models on the sentiment identification in terms of $R^2$ scores, averaged over 5 runs. `base' and `large' correspond to the number of steps performed to additionally pretrain base- or large-sized models, each row therefore requiring equal amount of time on a TPU. Step 0 in the columns for XB-BERTić, XL-BERTić and XL-SloBERTić represents the performance of the models prior to pretraining, i.e., the performance of the XLM-RoBERTa-base and XLM-RoBERTa-large. \label{parlasent}
}
\end{table*}

\subsubsection{Commonsense Reasoning}
\label{subsubsection:copa}
Tables \ref{res:copa-hr} and \ref{res:copa-sr} show the results of evaluation of the models on the task of commonsense reasoning on Croatian and Serbian COPA dataset respectively. We train and test each model ten times and report average accuracy scores.

\begin{table*}[!ht]
\begin{tabular}{r|r|rrrrr}
\hline
\bf base & \bf large & \bf cseBERT     & \bf BERTić      & \bf XB-BERTić & \bf XL-BERTić & \bf XL-SloBERTić \\
\hline
0    & 0     & 0.645±0.024 & 0.669±0.016 & 0.585±0.018 & 0.571±0.029 & 0.571±0.029   \\
12   & 6     &            &            & 0.602±0.021 & 0.651±0.025 & 0.616±0.018   \\
24   & 12    &            &            & 0.607±0.015 & 0.640±0.036 & 0.643±0.030   \\
36   & 18    &            &            & 0.585±0.019 & 0.656±0.026 & 0.654±0.027   \\
48   & 24    &            &            & 0.593±0.015 & 0.655±0.032 & 0.668±0.023   \\
60   & 30    &            &            & 0.589±0.023 & 0.658±0.033 & 0.641±0.020   \\
72   & 36    &            &            & 0.599±0.016 & 0.635±0.038 & 0.651±0.027   \\
84   & 42    &            &            & 0.604±0.024 & 0.644±0.034 & 0.656±0.033   \\
96   & 48    &            &            & 0.599±0.022 & 0.635±0.031 & 0.628±0.035  \\
\hline
\end{tabular}
\caption{Comparison of models on the commonsense reasoning on the Croatian COPA dataset in terms of accuracy scores, averaged over 10 runs. `base' and `large' correspond to the number of steps performed to additionally pretrain base- or large-sized models, each row therefore requiring equal amount of time on a TPU. Step 0 in the columns for XB-BERTić, XL-BERTić and XL-SloBERTić represents the performance of the models prior to pretraining, i.e., the performance of the XLM-RoBERTa-base and XLM-RoBERTa-large. \label{res:copa-hr}}
\end{table*}

\begin{table*}[!ht]
\begin{tabular}{r|r|rrrrr}
\hline
\bf base & \bf large & \bf cseBERT     & \bf BERTić      & \bf XB-BERTić & \bf XL-BERTić & \bf XL-SloBERTić \\
\hline
0    & 0     & 0.607±0.027 & 0.689±0.024 & 0.573±0.016 & 0.570±0.032 & 0.570±0.032   \\
12   & 6     &            &            & 0.605±0.016 & 0.642±0.022 & 0.613±0.021   \\
24   & 12    &            &            & 0.603±0.018 & 0.668±0.033 & 0.639±0.017   \\
36   & 18    &            &            & 0.598±0.030 & 0.685±0.034 & 0.650±0.022   \\
48   & 24    &            &            & 0.621±0.015 & 0.659±0.035 & 0.667±0.023   \\
60   & 30    &            &            & 0.609±0.032 & 0.640±0.030 & 0.649±0.030   \\
72   & 36    &            &            & 0.618±0.024 & 0.629±0.035 & 0.632±0.028   \\
84   & 42    &            &            & 0.628±0.024 & 0.630±0.036 & 0.666±0.031   \\
96   & 48    &            &            & 0.617±0.025 & 0.637±0.021 & 0.655±0.026  \\
\hline
\end{tabular}
\caption{Comparison of models on the commonsense reasoning on the Serbian COPA dataset in terms of accuracy scores, averaged over 10 runs. `base' and `large' correspond to the number of steps performed to additionally pretrain base- or large-sized models, each row therefore requiring equal amount of time on a TPU. Step 0 in the columns for XB-BERTić, XL-BERTić and XL-SloBERTić represents the performance of the models prior to pretraining, i.e., the performance of the XLM-RoBERTa-base and XLM-RoBERTa-large. \label{res:copa-sr}}
\end{table*}

\end{document}